%% file: 3DShoe.tex
\documentclass[conference]{IEEEtran}
\IEEEoverridecommandlockouts
\usepackage{cite}
\usepackage{amsmath,amssymb,amsfonts}
\usepackage{float}
\usepackage{algorithmic}
\usepackage{graphicx}
\usepackage{textcomp}
\usepackage{xcolor}
\def\BibTeX{{\rm B\kern-.05em{\sc i\kern-.025em b}\kern-.08em
    T\kern-.1667em\lower.7ex\hbox{E}\kern-.125emX}}
\begin{document}

\title{3D Reconstruction of Shoes for Augmented Reality\\
}

\author{
\IEEEauthorblockN{1\textsuperscript{st} Pratik Shrestha\IEEEauthorrefmark{1}, 
2\textsuperscript{nd} Sujan Kapali\IEEEauthorrefmark{1}, 
3\textsuperscript{rd} Swikar Gautam\IEEEauthorrefmark{1}, 
4\textsuperscript{th} Vishal Pokharel\IEEEauthorrefmark{1}, 
5\textsuperscript{th} Santosh Giri}
\IEEEauthorblockA{\textit{Department of Electronics and Computer Engineering} \\
\textit{Pulchowk Campus, Institute of Engineering}\\
Lalitpur, Nepal \\
\{077bct059.pratik, 077bct086.sujan, 077bct091.swikar, 077bct094.vishal, santoshgiri\}@pcampus.edu.np}
\IEEEauthorblockA{\IEEEauthorrefmark{1}These authors contributed equally to this work.}
}

\maketitle

\begin{abstract}
This paper introduces a mobile-based solution that enhances online shoe shopping through 3D modeling and Augmented Reality (AR), leveraging the efficiency of 3D Gaussian Splatting. Addressing the limitations of static 2D images, the framework generates realistic 3D shoe models from 2D images, achieving an average Peak Signal-to-Noise Ratio (PSNR) of 32, and enables immersive AR interactions via smartphones. A custom shoe segmentation dataset of 3120 images was created, with the best-performing segmentation model achieving an Intersection over Union (IoU) score of 0.95. This paper demonstrates the potential of 3D modeling and AR to revolutionize online shopping by offering realistic virtual interactions, with applicability across broader fashion categories. 
\end{abstract}

\begin{IEEEkeywords}
Augmented Reality (AR), 3D Modeling, Gaussian Splatting, Segmentation
\end{IEEEkeywords}

\section{Introduction}
The rapid advancements in 3D modeling and computer graphics have revolutionized various industries, enabling the creation of immersive and realistic Augmented Reality (AR) solutions. These technologies are increasingly utilized in the fashion industry to enhance customer experiences by enabling interaction and visualization of products before purchase. Despite these advancements, traditional online shopping platforms remain limited by their reliance on static 2D images, which fail to replicate the exploratory and sensory experience of physical retail environments.

Historically, generating 3D models from 2D images has been a labor-intensive and time-consuming process, requiring significant human intervention. However, recent breakthroughs have accelerated the modeling process while achieving higher levels of precision and realism. Among the approaches to 3D modeling, Photogrammetry, Neural Radiance Fields (NeRF), and Gaussian Splatting have emerged as prominent techniques, each with distinct advantages and limitations.

Photogrammetry, although widely used, encounters challenges such as handling reflective surfaces, dealing with occlusions, and requiring high-quality input images, which limit its practicality. NeRF, a deep learning-based method, generates highly detailed models but is computationally intensive, making real-time applications infeasible. In contrast, Gaussian Splatting offers a significant advantage by enabling faster training, real-time rendering, and easy modification of the generated models. This paper seeks to find techniques to leverage 3D Gaussian Splatting to streamline the creation of realistic 3D models from 2D images, focusing on its application in enhancing online shopping experiences.

\section{Literature Review}
Recent advancements in deep learning have revolutionized the synthesis and rendering of 3D models from 2D images, presenting efficient alternatives to conventional Photogrammetry techniques \cite{nerf}\cite{ngp}\cite{gaussian}. Among these advancements, two prominent methods, Neural Radiance Fields (NeRF) \cite{nerf} and 3D Gaussian Splatting \cite{gaussian}, have garnered considerable attention for their respective strengths and applications.

NeRF, pioneered by Shrinivasan, Mildenhall, and Tancik in 2020, operates by harnessing radiance fields for view synthesis. By employing Multi-Layer Perceptrons (MLPs), this method learns to predict both the volume density and color of a point based on a 5D input—comprising the spatial location of the camera and its viewing direction. Subsequently, NeRF synthesizes views by querying points along marching rays for color and volume density and applies classical volumetric rendering techniques for rendering. This technique excels in producing true-to-life renderings by learning the volumetric representation of a scene from a collection of images. However, its training process demands significant computational resources, making it impractical for real-time applications.

In contrast, 3D Gaussian Splatting, introduced by Kerbel, Kopanas, Leimkuhler, and Drettakis, leverages Gaussians to render views. Firstly, a point cloud is generated from images using the Structure from Motion(SfM). \cite{sfm} This approach involves the conversion of individual points within a point cloud into Gaussians, and fine-tuning the parameters of these Gaussians through stochastic gradient descent. Notably, this technique offers a substantial leap in speed compared to the original NeRF, enabling real-time rendering capabilities. As a result, it emerges as a promising solution for scenarios requiring instantaneous interactions, making it particularly suitable for applications such as virtual try-on experiences.

The image given to model for 3D reconstruction should have its background removed. There are wide variety of segmentation models\cite{unet}\cite{attention-unet}\cite{nnunet}\cite{sam}\cite{detic} available for this task. However, YOLOv8\cite{yolo}, leveraging state-of-the-art progressions in deep learning and computer vision, is commonly employed because of its light weight and development tools. YOLOv8 is proficient in various vision AI tasks, encompassing detection, segmentation, pose estimation, tracking, and classification.

Training a machine learning model for segmentation tasks typically demands a large, annotated dataset, which is both time-intensive and prone to human error. To address this challenge, generalized large-scale models like SAM \cite{sam} can be leveraged to automatically generate segmentation masks, significantly reducing the annotation effort.

Although Gaussian techniques yield state-of-the-art (SOTA) results, existing infrastructure, such as 3D modeling and editing software, as well as platforms offering augmented reality solutions, primarily operate with meshes. Consequently, SuGaR\cite{sugar} is employed for mesh extraction from 3D Gaussian splatting. Gaussian Splatting accelerates training for rendering intricate scenes but can result in disorganized Gaussians, complicating mesh creation. SuGaR addresses this by aligning Gaussians to the scene's surface, enhancing mesh generation through efficient Poisson reconstruction, which preserves details better than alternatives like Marching Cubes with Neural SDFs.

In 2021, AR-Shoe \cite{arshoe} was introduced, which leverages deep learning methodologies to superimpose a 3D shoe model onto an image of a foot. The model operates by taking foot images as inputs and producing keypoint heatmaps, part affinity fields maps (pafmaps), and segmentation maps of the feet. These outputs are utilized to derive the precise 6 degrees of freedom (6-DoF) pose for the feet, and overlay of 3D shoe which is then occluded in the input image.

Song et al. introduced VTONShoes\cite{vton}, an advanced AR-based real-time virtual shoe try-on system that achieves precise 6-DoF pose estimation and realistic rendering using dense keypoints, joint keypoint localization, and silhouette segmentation. The system's smooth and stable performance at 25-45 FPS, coupled with the introduction of the Diverse-Shoes dataset, marks a significant advancement in real-time AR virtual try-on technology.

\section{Methodology}
\subsection{Data Collection}\label{AA}
In the 3D Gaussian Splatting Model, we achieve accurate modeling by over-fitting a single object. Consequently, a vast data-set isn't necessary. But, we do need data for segmentation of shoes from the given images. For this purpose, we collected 101 videos, each lasting 30 seconds, from students at Pulchowk Campus. These videos were taken with different cameras, such as Poco X3, Samsung Galaxy S9, Redmi 12, and others.\\
From each video, we extracted 30 images by evenly dividing the video frames. These images were then used for shoe segmentation to generate 3D models. The segmentation process involved isolating the shoe from the background in each image. These segmented images formed the foundational data for our subsequent 3D modeling.
\begin{figure}[htbp]
    \centering
    \includegraphics[width=0.5\textwidth]{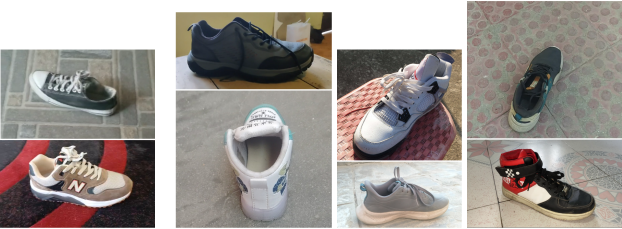}
    \caption{Sample Images from the Dataset}
    \label{fig: Sample Images from the Dataset}
\end{figure}

\subsection{Background Masking}
We used Meta's Segment Anything Model (SAM)\cite{sam} to generate segmentation maps from images, which provided accurate results. However, the model required over 6GB of GPU memory, making it costly to host on a server due to its high computational demands. To address this, we used SAM to create a dataset and then trained a smaller, more efficient model that significantly reduces computational requirements. The steps we followed are outlined below:

\begin{itemize}
    \item Dataset Collection: We collected 101 videos of distinct shoes each of length 30 seconds. We collected 30 images from each video by sampling uniformly across the duration of the video. This gave us a total of 3030 images.
    
    \item Annotation: We use an automated pipeline to annotate the images. First, we used a dataset\cite{boston-shoe} from Roboflow to train an YOLO model to detect shoes in images. Then, we used SAM to annotate the images, sending in the bounding box obtained from the YOLO model as prompt.
    \item Correction of improper annotations: The automated annotation pipeline did not provide correct annotation on all images which arose from inaccuracies in YOLO model while detecting shoes and inaccuracies in segmentation map from SAM. However, such instances were very few in number and hence, did not require much time to correct manually. We manually skimmed through all the annotations correcting the ones with errors to ensure the correctness of whole dataset. Furthermore, there were also some frames which did not have shoes. We removed such images.
    \item Training: For training, we split the data into train, validation and test set with 80\%, 10\% and 10\% of the total data respectively. We wrote the split script to ensure that images from a single video does not end up in multiple splits which might bump the accuracy simply by overfitting the training samples. We trained the dataset on YOLO v8 and Unet model pre-trained on the COCO128 dataset. The model provided results which is decent enough for the task of 3D reconstruction. The final model size is 6.5 MB which is orders of magnitude smaller than the SAM model which is about 1.5GB. This significantly reduces the required computational resources and server costs for hosting the model.
    \begin{figure}[htbp]
                    \includegraphics[width=0.2\textwidth]{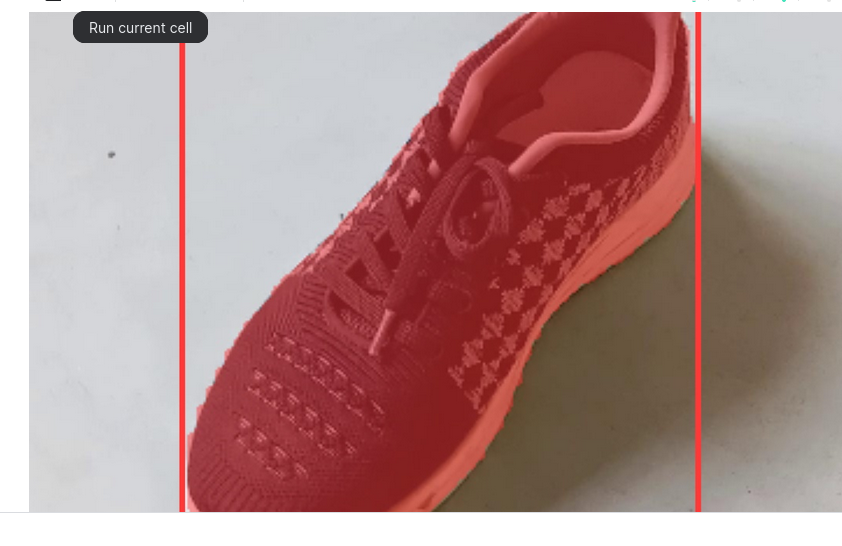}
                    \includegraphics[width=0.2\textwidth]{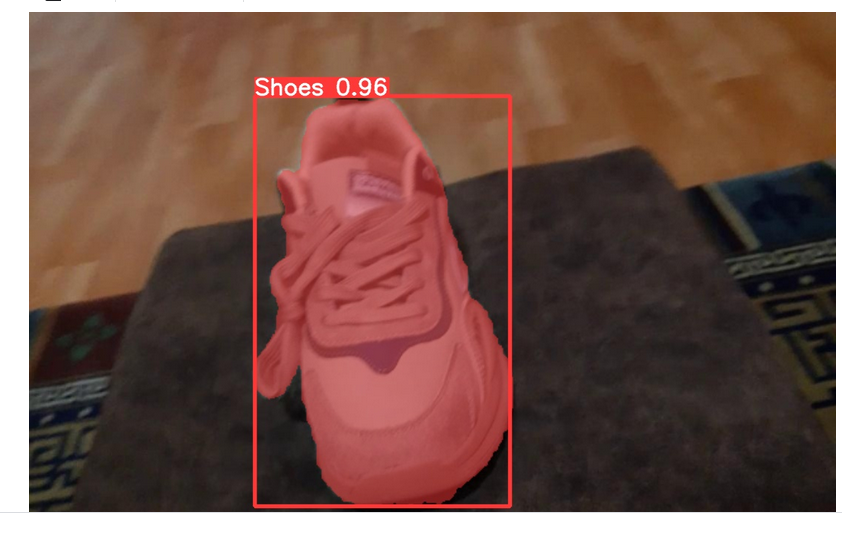}
                \caption{Prediction on Test set from trained YOLO v8 model}
                \label{fig: Prediction on Test set from YOLO model}
            \end{figure}
    
\end{itemize}

\subsection{Colmap}
We used COLMAP, a popular tool for 3D reconstruction, to process our data. It helped us create accurate 3D models from input images by estimating camera poses and generating sparse and dense point clouds. The images were first aligned using COLMAP's feature-matching and structure-from-motion pipeline. After alignment, dense reconstruction was performed to generate detailed 3D points. The output was then used as input for the next stages of our project.

\subsection{Gaussian Splatting Model}
We used the implementation provided in the Gaussian Splatting repository with some adjustments. The pre-processing step included cropping each image to its maximum size along both dimensions. The model was then trained for 7000 iterations, which took approximately 10 minutes.

\subsection{Mesh Extraction}
We used the implementation from the Sugar repository with some adjustments. The model was trained for 9000 refinement iterations, which was the most time-consuming step. Extracting the mesh took approximately an hour. The final images from the undistortion phase had a black background, resulting in many black faces in the final mesh. To address this, we wrote a script to remove all black vertices and extract the largest connected component, ensuring a cleaner mesh.
\begin{figure}[htbp]
        \centering
        \includegraphics[scale=0.3]{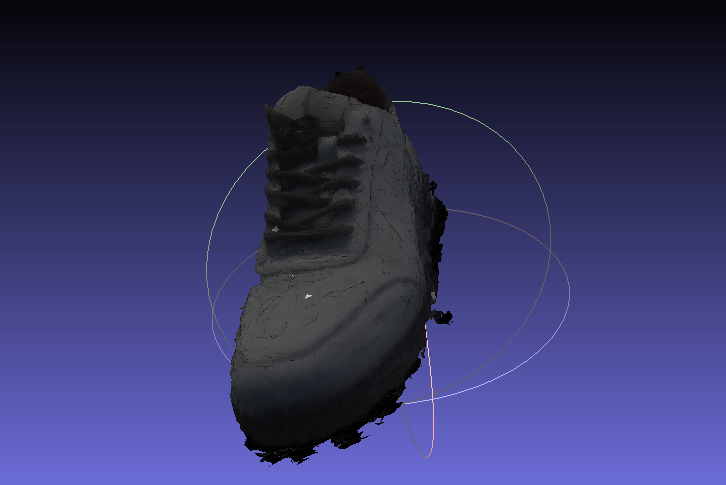}
    \caption{Extracted Mesh}
    \label{fig: Extracted Mesh}
\end{figure}

\subsection{Augmented Reality}
To predict the pose of the user’s feet in augmented reality (AR), we employed Lens Studio. The process of fitting a shoe onto the user’s leg involves the following steps:
    
    \begin{itemize}
        \item We began by creating a 3D model using Gaussian Splatting, then combined the obj, mtl, and png files using Blender.
        
        \item Next, we imported the GLB files of both the left and right foot shoes into Lens Studio.
    
        \item Using Lens Studio's leg detection template, we aligned the 3D model feet with the user's actual feet.
    
        \item To address occlusion, we placed transparent cylinders through the holes of the shoes.
    
        \item Finally, we published the lens in Lens Studio and integrated it into our Flutter app.
    
    \end{itemize}

      \begin{figure}[htbp]
                    \centering
                    \includegraphics[scale=0.5]{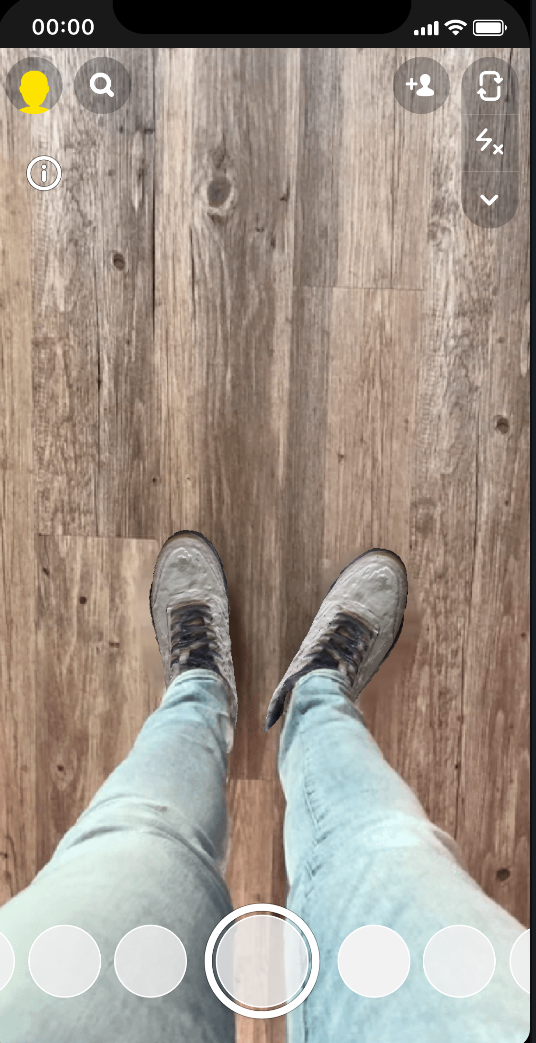}
                \caption{Results in AR}
                \label{fig: Results in AR}
        \end{figure}

\subsection{Mobile Application}
 We integrated individual pieces in a mobile application which allows user to enjoy AR experience in their mobile phones. We utilized flutter along with SnapAR package to develop the app.
     \begin{figure}[htbp]
          \centering
          \begin{minipage}{0.2\textwidth}
            \centering
            \includegraphics[scale=0.04]{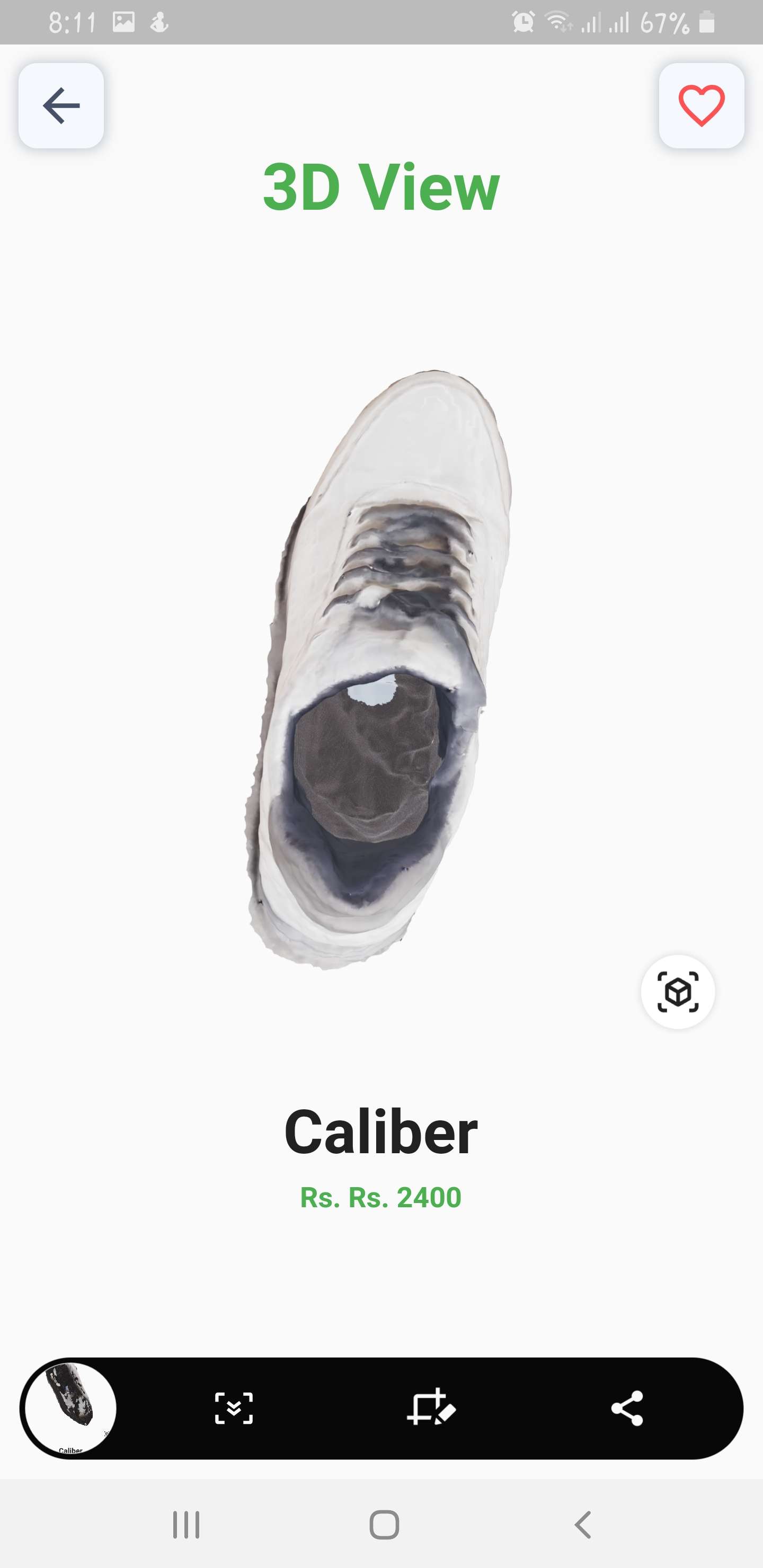}
          \end{minipage}
          \hfill
          \begin{minipage}{0.2\textwidth}
            \centering
            \includegraphics[scale=0.04]{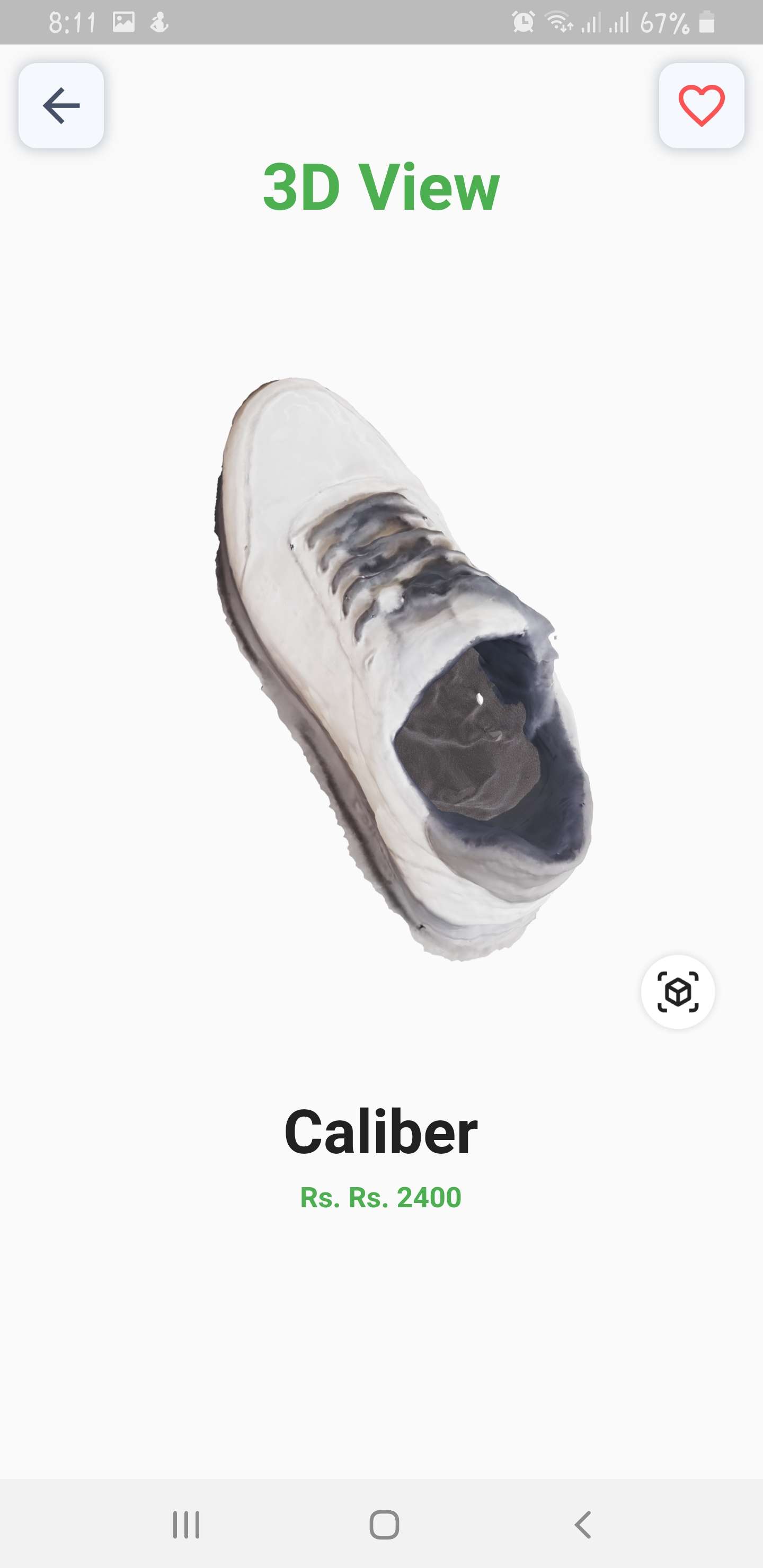}
          \end{minipage}
          \hfill
          \begin{minipage}{0.2\textwidth}
            \centering
            \includegraphics[scale=0.04]{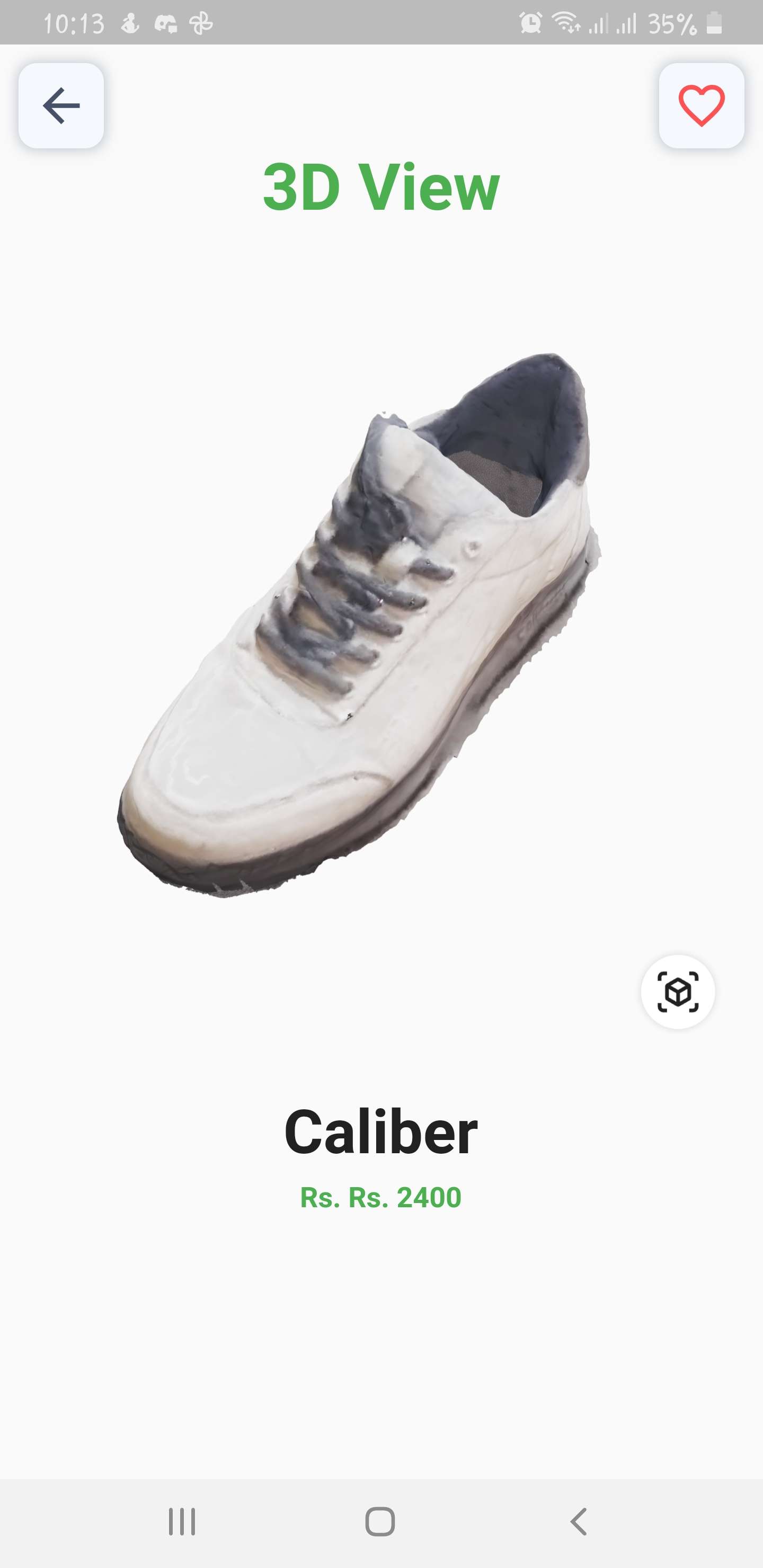}
          \end{minipage}
          \hfill
          \begin{minipage}{0.2\textwidth}
            \centering
            \includegraphics[scale=0.04]{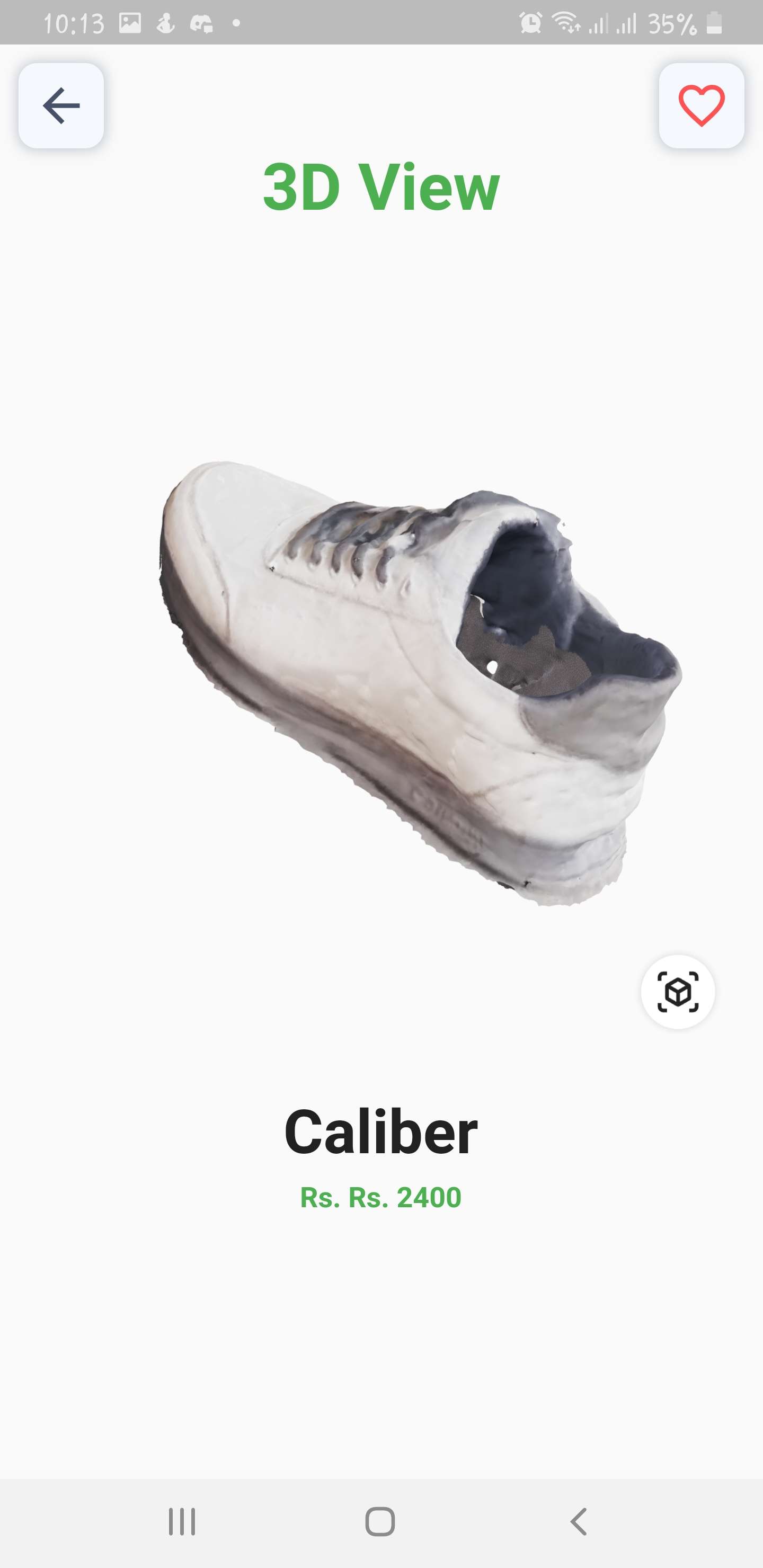}
          \end{minipage}
          \caption{Interactive 3D Viewer}
        \end{figure}

\subsection{Overall System Design}
The pipeline comprises several integral components working in tandem to create a comprehensive and immersive AR experience. First, the Background Masking element eliminates image backgrounds, facilitating the generation of masked images fed into the model for 3D shoe modeling. Colmap generates a 3D point cloud or mesh representing reconstructed scenes. The Gaussian splatting model refines this representation by using multiple 3D Gaussian distributions to create a smooth point cloud depiction. Sugar converts the obtain point cloud to mesh format. Foot pose estimation model involves keypoints prediction, pose estimation, and segmentation for occlusion identification, ensuring a realistic representation of the shoe and leg interaction. We utilize SnapChat's Camerakit to implement this functionality. The project culminates in a mobile application merging these components, enabling users to engage in AR.

\begin{figure}[htbp]
    \centering
    \includegraphics[scale=0.3]{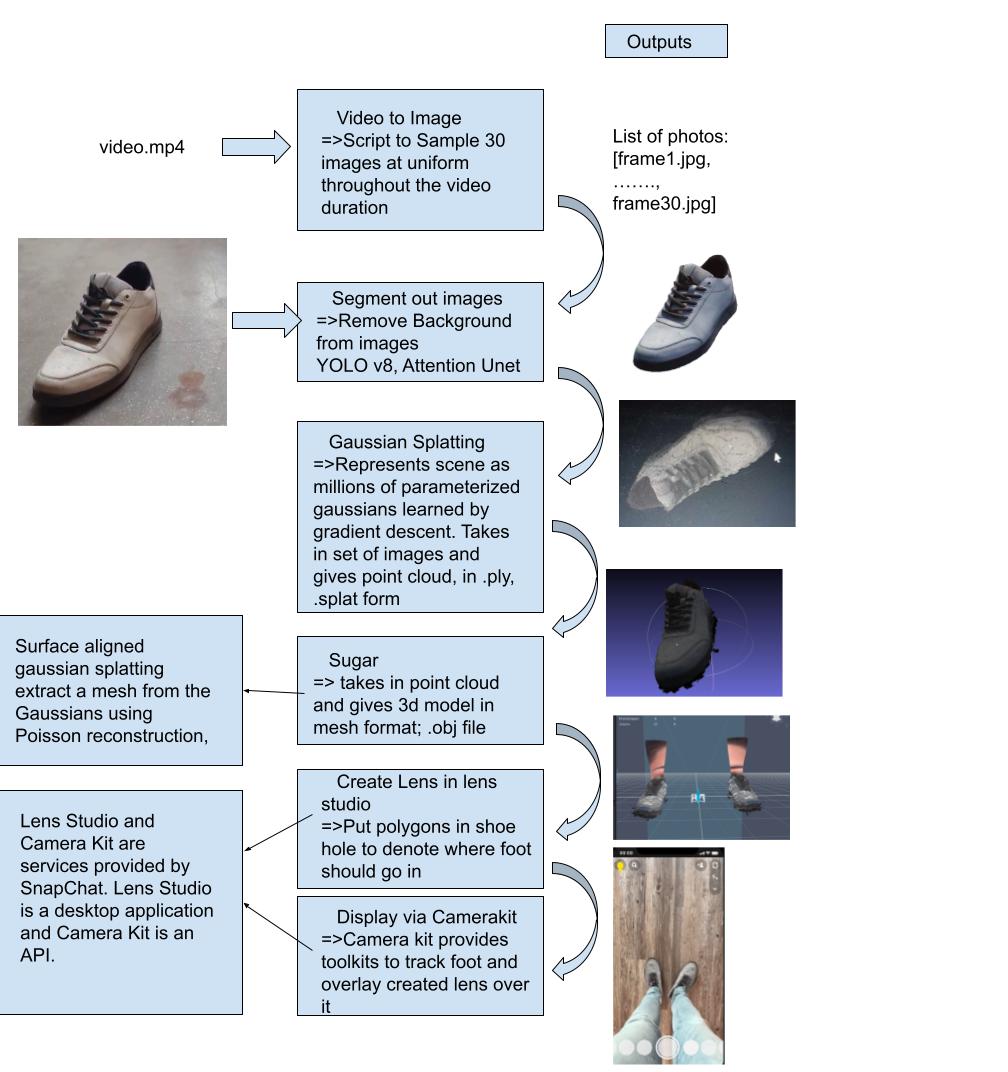}
    \caption{Overview of Workflow}
    \label{fig: Overview of Workflow}
\end{figure}

\subsection{Evaluation and Testing}
The quality of the final augmented output largely depends on the quality of the 3d model generated, we have to ensure that the initial output is as realistic as possible. We used \textbf{Peak Signal-to-Noise Ratio } to evaluate the performance of our method. 
\\
\\
The PSNR is expressed in decibels (dB) and is calculated using the following formula:
\begin{align}
    \text{PSNR} &= 10 \cdot \log_{10}\left(\frac{\text{MAX}^2}{\text{MSE}}\right)
\end{align}

where:
\begin{align*}
    \text{MAX} &= \text{maximum possible pixel value of the image or video} \\
    & \text{(e.g., 255 for an 8-bit grayscale image)}\\
    \text{MSE} &= \text{Mean Squared Error between the original and the}\\
    & \text{reconstructed image}
\end{align*}

The Mean Squared Error (MSE) is calculated as follows:
\begin{equation}
    \text{MSE} = \frac{1}{N \times M} \sum_{i=1}^{N} \sum_{j=1}^{M} \left(\text{Original}(i,j) - \text{Reconstructed}(i,j)\right)^2
\end{equation}

where:
\begin{align*}
    N & = \text{number of rows in the image or video} \\
    M & = \text{number of columns in the image or video}
\end{align*}
\\
For the segmentation task, we evaluate our models over IOU.
\begin{equation}
IOU = \frac{Area of Intersection}{Area of Union} = \frac{TP}{TP + FP + FN}
\end{equation}
\begin{align*}
    \text{True Positive (TP)} & : \text{Instances where the model correctly predicts} \\ & \text{the presence of a positive class}\\
    \text{False Positive (FP)} & : \text{Instances where the model incorrectly predicts} \\& \text{the presence of a positive class}\\
    \text{False Negative (FN)} & : \text{Instances where the model fails to predict}\\& \text{the presence of a positive class}
\end{align*}
\\
The final qualitative evaluation of our application was done by testing the system on various input conditions. A robust system can  handle various conditions of different foot poses
and can realize a realistic AR effect in practical scenes.

\section{Results and Discussions}
Following are the results we obtained grouped by system components:
\\
\subsection{Background Masking}
The data collection phase resulted in 3,000 images. Pre-processing was done as explained in Implementation section. Following are the results we obtained after training different models. 
\\
   We trained the dataset on YOLO v8 and Unet model pre-trained on the COCO128 dataset. The model provided results which is decent enough for the task of 3D reconstruction. The final model size is 6.5 MB which is orders of magnitude smaller than the SAM model which is about 1.5GB. This significantly reduces the required computational resources and server costs for hosting the model.

    \begin{figure}[htbp]
            \centering
            \includegraphics[scale=0.2]{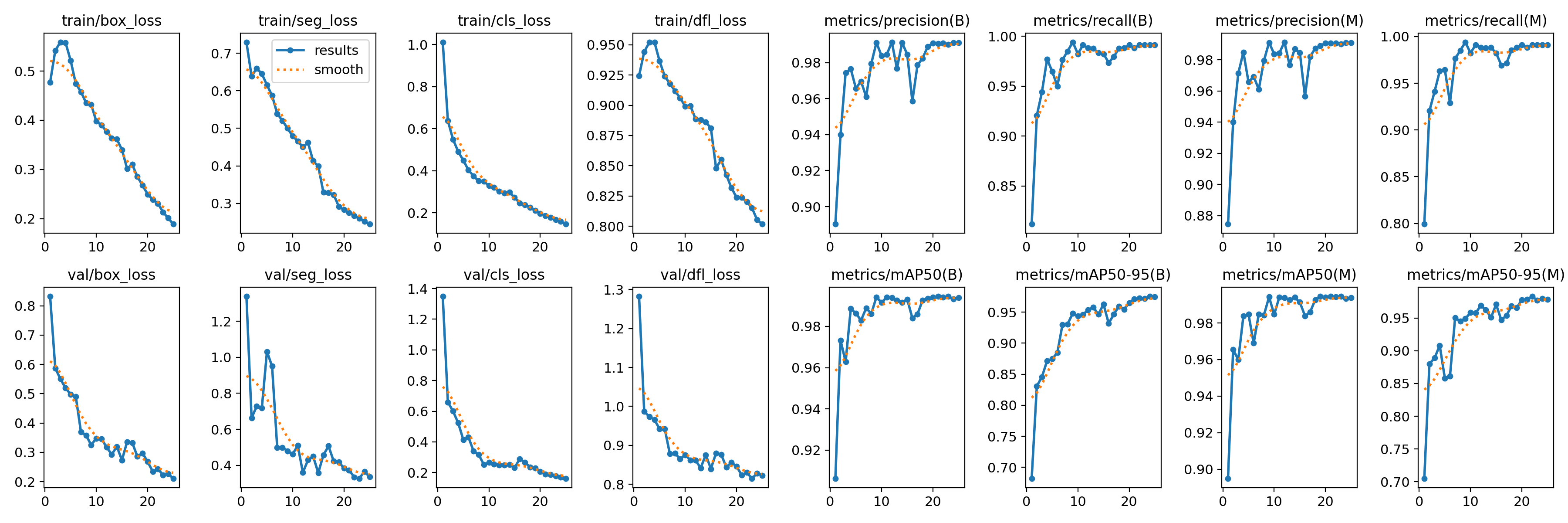}
        \caption{Training Graph of YOLO V8n}
        \label{fig: Training Graph of YOLO Vn}
    \end{figure}

    \begin{figure}[htbp]
        \centering
        \includegraphics[scale=0.4]{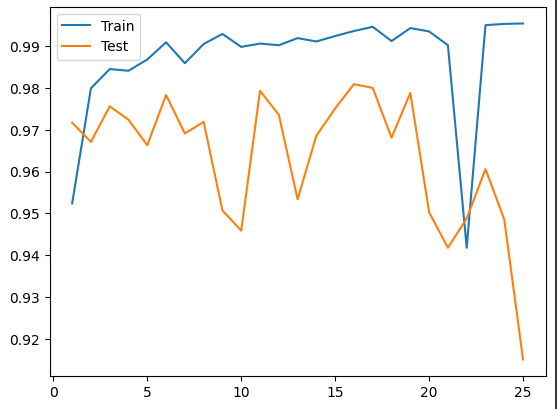}
        \caption{Training Graph of Attention Unet (IoU)}
        \label{fig: Training Graph of Attention Unet (IoU)}
    \end{figure}

\begin{table}[h]
\renewcommand{\tablename}{} 
\renewcommand{\thetable}{} 
\renewcommand{\arraystretch}{1.2} 

\centering
\begin{tabular}{| p{5em} | p{4em} | p{3em} | p{5em} |} 
  \hline
  \textbf{Model} & \textbf{Parameters} & \textbf{Epochs} & \textbf{IOU (threshold=0.5)} \\ 
  \hline
  YOLOv8n & 2.7M & 100 & 0.9494 \\ 
  \hline
  YOLOv8m & 27.3M & 25 & 0.9577 \\ 
  \hline
  Unet (ResNet) & 32M & 25 & 0.9548 \\ 
  \hline
\end{tabular}
\caption{Comparison of Segmentation Models (Test Set)}
\end{table}

\noindent In all the models IoU spiked during the initial epoch and saturated quickly. The training loss kept on decreasing, but the training was stopped after validation loss started to saturate. Among the models, YOLOv8m performed the best on the test set. However, it should be noted that each model was trained for an hour. Results may vary if training is left to progress for more time.

\subsection{Gaussian Splatting Model}
We drew random samples from our dataset and evaluated the Gaussian Splatting model. We obtained an average PSNR of 34.
    \begin{figure}[htbp]
            \centering
            \includegraphics[scale=0.4]{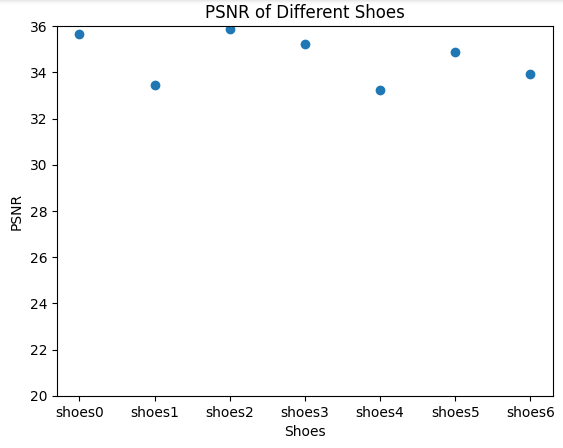}
        \caption{PSNR of different shoes}
        \label{fig: PSNR of different shoes}
    \end{figure}
\\
\\
\subsection{Mesh Extraction}

The result of Gaussian splatting was used to extract mesh using Sugar. Following is the trajectory of loss we obtained for a shoe:
\begin{figure}[htbp]
        \centering
        \includegraphics[scale=0.4]{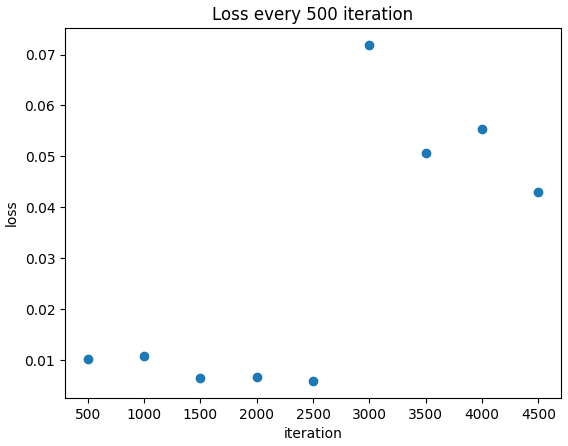}
    \caption{Loss every 500 iteration}
    \label{fig: Loss every 500 iteration}
\end{figure}
Initially the loss decreased, but after some iterations it abruptly increased. For optimal results, best model should be saved.

\section{Conclusion}
This work presents a robust system for 3D shoe modeling and AR integration, demonstrating the potential of Gaussian Splatting for efficient and accurate 3D reconstruction. By addressing limitations in data processing and model complexity, the proposed framework achieves realistic rendering suitable for mobile applications. Future efforts will focus on enhancing real-time rendering capabilities, improving occlusion handling, and extending the pipeline to other product categories for broader application in the fashion industry.

\section*{Acknowledgment}

We extend our heartfelt gratitude to Asst. Prof. Santosh Giri and Asst. Prof. Bibha Sthapit for their exceptional guidance and encouragement throughout this project. We also thank Assoc. Prof. Dr. Jyoti Tandukar for providing valuable feedback.

Our sincere appreciation goes to the Department of Electronics and Computer Engineering, Pulchowk Campus, for granting us the opportunity and offering consistent support and supervision.

Lastly, we are grateful to our colleagues and peers for their contributions, feedback, and collaboration, which have been vital to the success of this project.

\bibliographystyle{plain}
\input{3DShoe.bbl} 

\vspace{12pt}
\end{document}

%% file: 3DShoe.bbl

%% file: 3DShoe.bbl
\begin{thebibliography}{10}
\providecommand{\url}[1]{#1}
\csname url@samestyle\endcsname
\providecommand{\newblock}{\relax}
\providecommand{\bibinfo}[2]{#2}
\providecommand{\BIBentrySTDinterwordspacing}{\spaceskip=0pt\relax}
\providecommand{\BIBentryALTinterwordstretchfactor}{4}
\providecommand{\BIBentryALTinterwordspacing}{\spaceskip=\fontdimen2\font plus
\BIBentryALTinterwordstretchfactor\fontdimen3\font minus \fontdimen4\font\relax}
\providecommand{\BIBforeignlanguage}[2]{{%
\expandafter\ifx\csname l@#1\endcsname\relax
\typeout{** WARNING: IEEEtran.bst: No hyphenation pattern has been}%
\typeout{** loaded for the language `#1'. Using the pattern for}%
\typeout{** the default language instead.}%
\else
\language=\csname l@#1\endcsname
\fi
#2}}
\providecommand{\BIBdecl}{\relax}
\BIBdecl

\bibitem{nerf}
\BIBentryALTinterwordspacing
B.~Mildenhall, P.~P. Srinivasan, M.~Tancik, J.~T. Barron, R.~Ramamoorthi, and R.~Ng, ``Nerf: Representing scenes as neural radiance fields for view synthesis,'' \emph{CoRR}, vol. abs/2003.08934, 2020. [Online]. Available: \url{https://arxiv.org/abs/2003.08934}
\BIBentrySTDinterwordspacing

\bibitem{ngp}
\BIBentryALTinterwordspacing
T.~M\"uller, A.~Evans, C.~Schied, and A.~Keller, ``Instant neural graphics primitives with a multiresolution hash encoding,'' \emph{ACM Trans. Graph.}, vol.~41, no.~4, pp. 102:1--102:15, Jul. 2022. [Online]. Available: \url{https://doi.org/10.1145/3528223.3530127}
\BIBentrySTDinterwordspacing

\bibitem{gaussian}
B.~Kerbl, G.~Kopanas, T.~Leimkuhler, and G.~Drettakis, ``3d gaussian splatting for real-time radiance field rendering,'' \emph{ACM Journals}, vol.~42, no.~4, pp. 1--14, 2023.

\bibitem{sfm}
J.~L. Schönberger and J.-M. Frahm, ``Structure-from-motion revisited,'' \emph{IEEE}, 2016.

\bibitem{unet}
\BIBentryALTinterwordspacing
O.~Ronneberger, P.~Fischer, and T.~Brox, ``U-net: Convolutional networks for biomedical image segmentation,'' \emph{CoRR}, vol. abs/1505.04597, 2015. [Online]. Available: \url{http://arxiv.org/abs/1505.04597}
\BIBentrySTDinterwordspacing

\bibitem{attention-unet}
\BIBentryALTinterwordspacing
O.~Oktay, J.~Schlemper, L.~L. Folgoc, M.~C.~H. Lee, M.~P. Heinrich, K.~Misawa, K.~Mori, S.~G. McDonagh, N.~Y. Hammerla, B.~Kainz, B.~Glocker, and D.~Rueckert, ``Attention u-net: Learning where to look for the pancreas,'' \emph{CoRR}, vol. abs/1804.03999, 2018. [Online]. Available: \url{http://arxiv.org/abs/1804.03999}
\BIBentrySTDinterwordspacing

\bibitem{nnunet}
\BIBentryALTinterwordspacing
F.~Isensee, J.~Petersen, A.~Klein, D.~Zimmerer, P.~F. Jaeger, S.~Kohl, J.~Wasserthal, G.~K{\"{o}}hler, T.~Norajitra, S.~J. Wirkert, and K.~H. Maier{-}Hein, ``nnu-net: Self-adapting framework for u-net-based medical image segmentation,'' \emph{CoRR}, vol. abs/1809.10486, 2018. [Online]. Available: \url{http://arxiv.org/abs/1809.10486}
\BIBentrySTDinterwordspacing

\bibitem{sam}
A.~Kirillov, E.~Mintun, N.~Ravi, H.~Mao, C.~Rolland, L.~Gustafson, T.~Xiao, S.~Whitehead, A.~C. Berg, W.-Y. Lo, P.~Dollár, and R.~Girshick, ``Segment anything,'' 2023.

\bibitem{detic}
\BIBentryALTinterwordspacing
X.~Zhou, R.~Girdhar, A.~Joulin, P.~Kr{\"{a}}henb{\"{u}}hl, and I.~Misra, ``Detecting twenty-thousand classes using image-level supervision,'' \emph{CoRR}, vol. abs/2201.02605, 2022. [Online]. Available: \url{https://arxiv.org/abs/2201.02605}
\BIBentrySTDinterwordspacing

\bibitem{yolo}
\BIBentryALTinterwordspacing
Ultralytics. (2024) Yolov8. [Online]. Available: \url{https://github.com/ultralytics/ultralytics}
\BIBentrySTDinterwordspacing

\bibitem{sugar}
A.~Gu{\'e}don and V.~Lepetit, ``Sugar: Surface-aligned gaussian splatting for efficient 3d mesh reconstruction and high-quality mesh rendering,'' \emph{arXiv preprint arXiv:2311.12775}, 2023.

\bibitem{arshoe}
\BIBentryALTinterwordspacing
S.~An, G.~Che, J.~Guo, H.~Zhu, J.~Ye, F.~Zhou, Z.~Zhu, D.~Wei, A.~Liu, and W.~Zhang, ``Arshoe: Real-time augmented reality shoe try-on system on smartphones,'' \emph{CoRR}, vol. abs/2108.10515, 2021. [Online]. Available: \url{https://arxiv.org/abs/2108.10515}
\BIBentrySTDinterwordspacing

\bibitem{vton}
W.~Song, Y.~Gong, and Y.~Wang, ``Vtonshoes: Virtual try-on of shoes in augmented reality on a mobile device,'' in \emph{2022 IEEE International Symposium on Mixed and Augmented Reality (ISMAR)}, 2022, pp. 234--242.

\bibitem{boston-shoe}
\BIBentryALTinterwordspacing
Boston university cdr2g - shoes segmentation. [Online]. Available: \url{https://universe.roboflow.com/boston-university-cdr2g/shoes-seg}
\BIBentrySTDinterwordspacing

\end{thebibliography}
